\documentclass[conference]{IEEEtran}

\IEEEoverridecommandlockouts
\usepackage{cite}
\usepackage{amsmath,amssymb,amsfonts}
\usepackage{algorithmic}
\usepackage{graphicx}
\usepackage{textcomp}
\usepackage{xcolor}
\usepackage{url}
\def\BibTeX{{\rm B\kern-.05em{\sc i\kern-.025em b}\kern-.08em
    T\kern-.1667em\lower.7ex\hbox{E}\kern-.125emX}}

\begin{document}

\title{Cloud-based Digital Twin for Cognitive Robotics\\
\thanks{
This research was partially funded by the German Research Foundation DFG, as part of Collaborative Research Center (Sonderforschungsbereich) 1320 Project-ID 329551904 ``EASE - Everyday Activity Science and Engineering'', University of Bremen (\emph{http://www.ease-crc.org/}). 
The research reported in this paper has been also partially supported by the German Federal Ministry of Education and Research; Project-ID 16DHBKI047 "IntEL4CoRo", University of Bremen.}
}

\author{\IEEEauthorblockN{1\textsuperscript{st} Arthur Nied\'{zwiecki}}
\IEEEauthorblockA{\textit{Institute for Artificial Intelligence} \\
\textit{University of Bremen}\\
Bremen, Germany \\
0009-0009-8894-3241}
\and
\IEEEauthorblockN{2\textsuperscript{nd} Sascha Jongebloed}
\IEEEauthorblockA{\textit{Institute for Artificial Intelligence} \\
\textit{University of Bremen}\\
Bremen, Germany \\
0009-0006-3964-9597}
\and
\IEEEauthorblockN{3\textsuperscript{rd} Yanxiang Zhan}
\IEEEauthorblockA{\textit{Institute for Artificial Intelligence} \\
\textit{University of Bremen}\\
Bremen, Germany \\
0009-0007-3509-2661}
\and
\IEEEauthorblockN{4\textsuperscript{th} Michaela Kümpel}
\IEEEauthorblockA{\textit{Institute for Artificial Intelligence} \\
\textit{University of Bremen}\\
Bremen, Germany \\
0000-0002-0408-3953}
\and
\IEEEauthorblockN{5\textsuperscript{th} Jörn Syrbe}
\IEEEauthorblockA{\textit{Institute for Artificial Intelligence} \\
\textit{University of Bremen}\\
Bremen, Germany \\
0009-0001-2308-9517}
\and
\IEEEauthorblockN{6\textsuperscript{th} Michael Beetz}
\IEEEauthorblockA{\textit{Institute for Artificial Intelligence} \\
\textit{University of Bremen}\\
Bremen, Germany \\
0000-0002-7888-7444}
}

\maketitle


\begin{abstract}

    The paper presents a novel cloud-based digital twin learning platform for teaching and training concepts of cognitive robotics. Instead of forcing interested learners or students to install a new operating system and bulky, fragile software onto their personal laptops just to solve tutorials or coding assignments of a single lecture on robotics, it would be beneficial
    to avoid technical setups and directly dive into the content of cognitive robotics. 
    To achieve this, the authors utilize containerization technologies and Kubernetes to deploy and operate containerized applications, including robotics simulation environments and software collections based on the Robot operating System (ROS). The web-based Integrated Development Environment JupyterLab is integrated with RvizWeb and XPRA to provide real-time visualization of sensor data and robot behavior in a user-friendly environment for interacting with robotics software. The paper also discusses the application of the platform in teaching Knowledge Representation, Reasoning, Acquisition and Retrieval, and Task-Executives. The authors conclude that the proposed platform is a valuable tool for education and research in cognitive robotics, and that it has the potential to democratize access to these fields. The platform has already been successfully employed in various academic courses, demonstrating its effectiveness in fostering knowledge and skill development.
\end{abstract}


\section{Introduction}

    The cognitive robotics research field incorporates engineering, computer science, and psychology research to make a real robot move as intended. 
    Interpreting sensor data and actuating the motors in the intended way to successfully perform tasks in the real world is a complex problem that needs to be divided into multiple modules, for example, perception, motion planning, reasoning over knowledge, navigation systems, trajectory calculation, and task execution. There are expert researchers in all of these modules, and each researcher, depending on the field, has a different perspective on how to implement an intelligent robot. But to enable a cognitive holistic intelligent robot it is necessary to integrate all these fields.

    The complexity of such systems means that they are often difficult to penetrate and can only be operated by experts at great expense. The consequence of this is that there are only a few experts for cognitive robots, usually specialized only in certain hardware, simply because the often expensive hard- and software are not available to everyone. It is easy to imagine that as diverse the researchers and research fields are, teaching and training cognitive robotics is as complex as the diversity of the field and, therefore, includes various frameworks.
   
\begin{figure}[t]
    \centerline{\includegraphics[width=\linewidth]{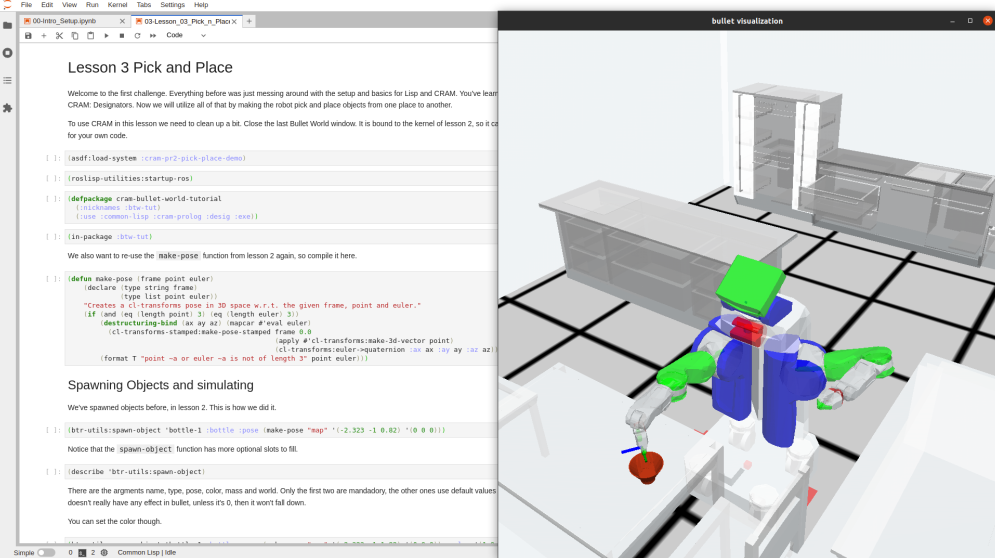}}
    \caption{Assignment on Pick\&Place robots on the left, physics simulation on the right, showing the robots believed world state.}
    \label{fig:jupyter}
\end{figure}

    We believe, to increase the number of people and collaboration in the field of cognitive robotics, such frameworks need to be made more accessible to research, industry and education. This includes lowering the inhibition threshold by raising less hardware and software requirements as well as providing intuitive tutorials that can be used by everybody.
    
    We propose to use a \emph{digital twin} learning platform to teach \& train concepts of cognitive robotic frameworks. A \emph{digital twin}, \emph{``the digital equivalent to a physical product''}~\cite{grieves2011virtually} provides a digital equivalent of a real laboratory, including a robot and its programmable frameworks. If such digital twin learning platforms are made available through a cloud, the requirements on hardware and software can be reduced tremendously, thus solving the aforementioned problems. A digital twin also allows for a visualization of the environments and robots. Such visualizations are essential on a robotics learning platform, as it provides immediate visual feedback and an intuitive and interactive way for learners to understand complex robotic behavior and environmental interaction.

    This article introduces a digital twin laboratory framework that can be accessed by researchers and students through the web service JupyterHub \cite{Jupyter}, a framework combining documentation and executable code for learning programming skills. 
    Through this programming interface, the learners interact with leading-edge research software, pre-installed alongside the Robot Operating System \cite{quigley2009ros} (ROS). 
    Through RVizWeb the programmer is able to visualize sensor data like camera images and laser scans in real-time. Jupyter, ROS, AI-applications, and RViz are wrapped in a Docker container, while Kubernetes \cite{Kubernetes} orchestrates these containers like classroom sessions for multiple students.



In section \ref{ch:related-work} we describe and discuss similiar frameworks to ours. The relevant terms and concepts are outlined in section \ref{cha:background}, while section \ref{cha:proposal_architecture} is dedicated to presenting our developed architecture. Section \ref{cha:application} focuses on validating our contributions by applying the developed platform in educational contexts and discussing the results. The paper concludes in section \ref{cha:conclusion}.

\section{Related Work}
\label{ch:related-work}


In the field of education and research, the utilization of platforms to increase the accessibility of robotics in teaching and training garnered considerable attention. One example for this are remote and virtual laboratory platforms.

Remote laboratory platforms like REAL\cite{guimaraes2003}, SyRoTek\cite{kulich2013} and "Robot Programming Network"\cite{casan2015} enable users to teleoperate robots either through specialized software or via web interfaces, facilitating a hands-on learning experience from a distance. Platforms like OpenUAV\cite{schmittle2018} or RobUALab\cite{jara2011} also offer a virtual laboratory with the capability to run simulations, either locally on the user's machine or remotely on a server. While this conventional teleoperation offers several advantages, our work focuses on simulations on the cloud, which reduces the limitations imposed by physical hardware availability and maintenance to enhance accessibility and scalability. Furthermore our platform integrates widely recognized physics simulations, such as PyBullet, ensuring a robust and realistic simulation environment.

Similar to our framework, the system developed by Avila et. al. also relies on ROS as its core robotics framework and is designed for instruction in AI software for robots\cite{avila2022}. Concurrently, Zhang et. al. have introduced an online platform that facilitates learning by allowing the execution of code through IPython kernels\cite{zhang2023dive}, an interactive Python interpreter. This platform, although restricted to Python's visualization capabilities,  demonstrates a platform similar to ours in educational settings. Lumpp et al. present a methodology for programming robotic tasks using Docker and Kubernetes, focusing on containerization and orchestration in ROS-based applications\cite{lumpp2021}. They successfully demonstrate the approach with a case study involving a Robotnik RB-Kairos robot in an industrial setting. 

The idea of using Digital Twins is well-established in the industry. For example, Intrinsics Flowstate \cite{Intrinsic} is using Digital Twins to develop concepts of industrial robots on a web-based construction platform with a graphical programming interface. 

Among existing platforms utilizing a digital-twin-like photo-realistic simulation, 'The Construct'\cite{tellez2016} stands out as the most similar to the in this article proposed framework, offering a cloud-based learning environment complete with visualization tools and simulation. However, contrary to 'The Construct', our platform is built on open-source software, providing unrestricted and free access to our educational \& training resources. This openness is a fundamental aspect of our approach aimed at democratizing learning in robotics. Furthermore, our approach exceeds the basics of ROS to teach holistic, cognition-enabled autonomous systems.

\section{Background}
\label{cha:background}

The following chapter describes the terms and concepts used to explain the proposed architecture.

\paragraph{Semantic Digital Twin}

In this study, we explore the application of \emph{semantic Digital Twins} (semDTs), which are semantically enhanced virtual models of real environments and their associated entities~\cite{Kuempel2021}, such as laboratories or production settings that can autonomously be created by robots as described in~\cite{Beetz2022}. Digital twins have garnered significant interest in the research community. Our focus is on leveraging these virtual representations, particularly of robotic laboratories, including the objects and robots within, for educational purposes, like the pick \& place lesson in figure \ref{fig:jupyter}. Such a setup democratizes access to advanced robotics education, making it more inclusive and far-reaching.



    \begin{figure}[t]
        \centerline{\includegraphics[width=1\linewidth]{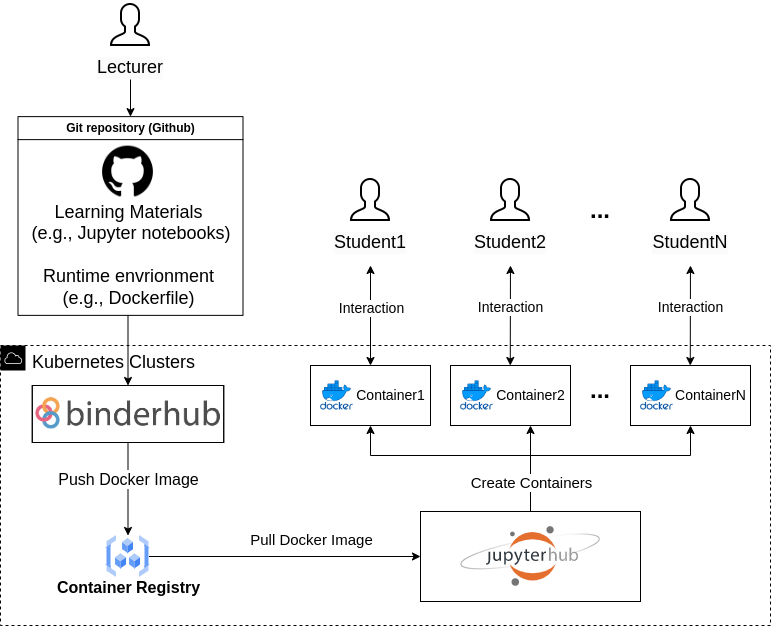}}
        \caption
        {The cloud service framework built on BinderHub \cite{ragan2018binder} initiates a new docker container for each connecting student. The Figure \ref{fig:arch-2} showcasing the robot software framework within the container.}
        \label{fig:arch}
    \end{figure}

\paragraph{Cognitive Robotics}

Concepts of cognitive robots are best described by Sandini, Vernon and Sciutti in \cite{Sandini2020}:\\
 "The word cognition derives from the Latin verb cognoscere, a composition of con (meaning related to) and gnoscere (to know). Cognitive robotics, then, is the branch of robotics where knowledge plays a central role in supporting action selection, execution, and understanding. [...]"\\
 The complexity of decision-making invites a broad field of research, from computer science over (electrical) engineering to psychology and philosophy.


\paragraph{Varying Environments and Robot Platforms}
Robots and their environments are versatile. Since a mobile robot can be applied in different environments, the \emph{digital twin} must be flexible too.
A robot focused on Human-Robot-Interaction (HRI) may be better suited for an everyday environment than a search-and-rescue robot. Dynamically exchanging the robot and its environment allows the researcher to tailor their experiments to their needs.

\paragraph{Cloud-Based}
Cloud-based services leverage remote computing resources and storage, offering significant reproducibility, flexibility, and scalability for education and research. The rapid development of AI technologies like machine learning, which require substantial computational resources, has given rise to open-source cloud service solutions in recent years. These state-of-the-art solutions are well-suited for building cloud-based robot simulation services that equally require heavy computational power. Educators, students, and researchers can rapidly build and test various robot algorithms and control methods without the need to establish and maintain complex simulation environments locally. In this study, we focus on offering a Platform-as-a-Service (PaaS)\cite{doelitzscher2011private} that allows educators and researchers to publish various Software-as-a-Service (SaaS), such as tutorials, coding assignments for courses, and robotics competition arenas.

\section{Proposed Architecture}
    \label{cha:proposal_architecture}
    
        The following chapter explains the components of the cloud-based \emph{digital twin} architecture.

    \begin{figure}[t]
        \centerline{\includegraphics[width=\linewidth]{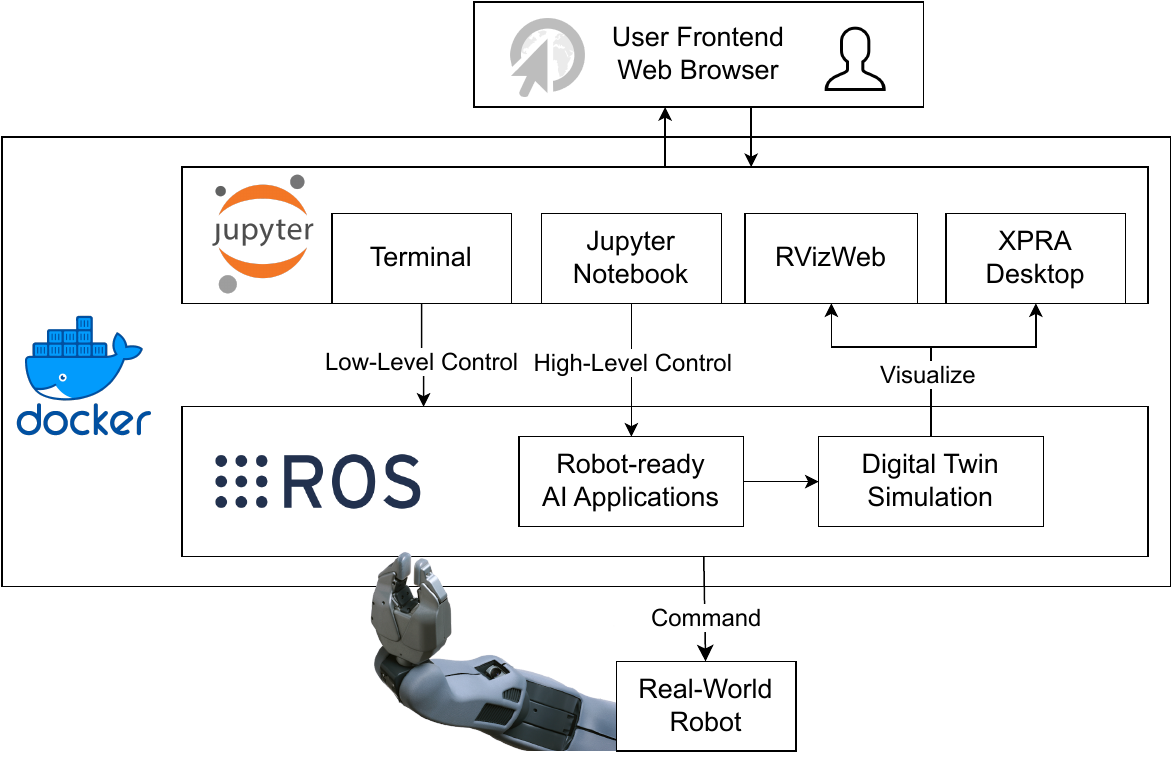}}
        \caption
        {To control a real-world robot, ROS gives commands to it. ROS is running in the Docker and is loaded with AI Applications, e.g. the Cognitive Robot Abstract Machine (CRAM). The Digital Twin Simulation, a 3D game engine, is visualized in Jupyter via RVizWeb and/or a lightweight display for visual applications. Jupyter Notebooks run code to control the AI Applications, while a Terminal allows for direct access to the ROS processes.}
        \label{fig:arch-2}
    \end{figure}

    \paragraph{Containerized Application}

        Containerized applications use containerization technologies to package the application software along with necessary runtime libraries and configuration files into one executable process. The presented approach uses Docker \cite{merkel2014docker}. This speeds up the deployment of software services across different computing environments without concerning the underlying hardware and operating systems. Each container is an isolated Linux environment, ensuring that each application functions without being affected by others, and behaves consistently.
    
        Many open-source robot research software, such as ROS and Gazebo, are developed based on Linux systems, leading to the widespread of containerization solutions in the field of robotics research. Figure \ref{fig:arch-2} depicts a containerized application we constructed for robotics learning. It contains Robot Operating System (ROS)\cite{quigley2009ros}, our research software, an web Integrated Development Environment (JupyterLab \cite{perkel2018jupyter}) and several visualization tools (RvizWeb, XPRA desktop). 
        With this containerized application, students can launch a full-stack learning environment on their personal PC quickly and control robots within the Local-Area-Network through the web-based interface.
    

\begin{figure}[t]
    \centerline{\includegraphics[width=\linewidth]{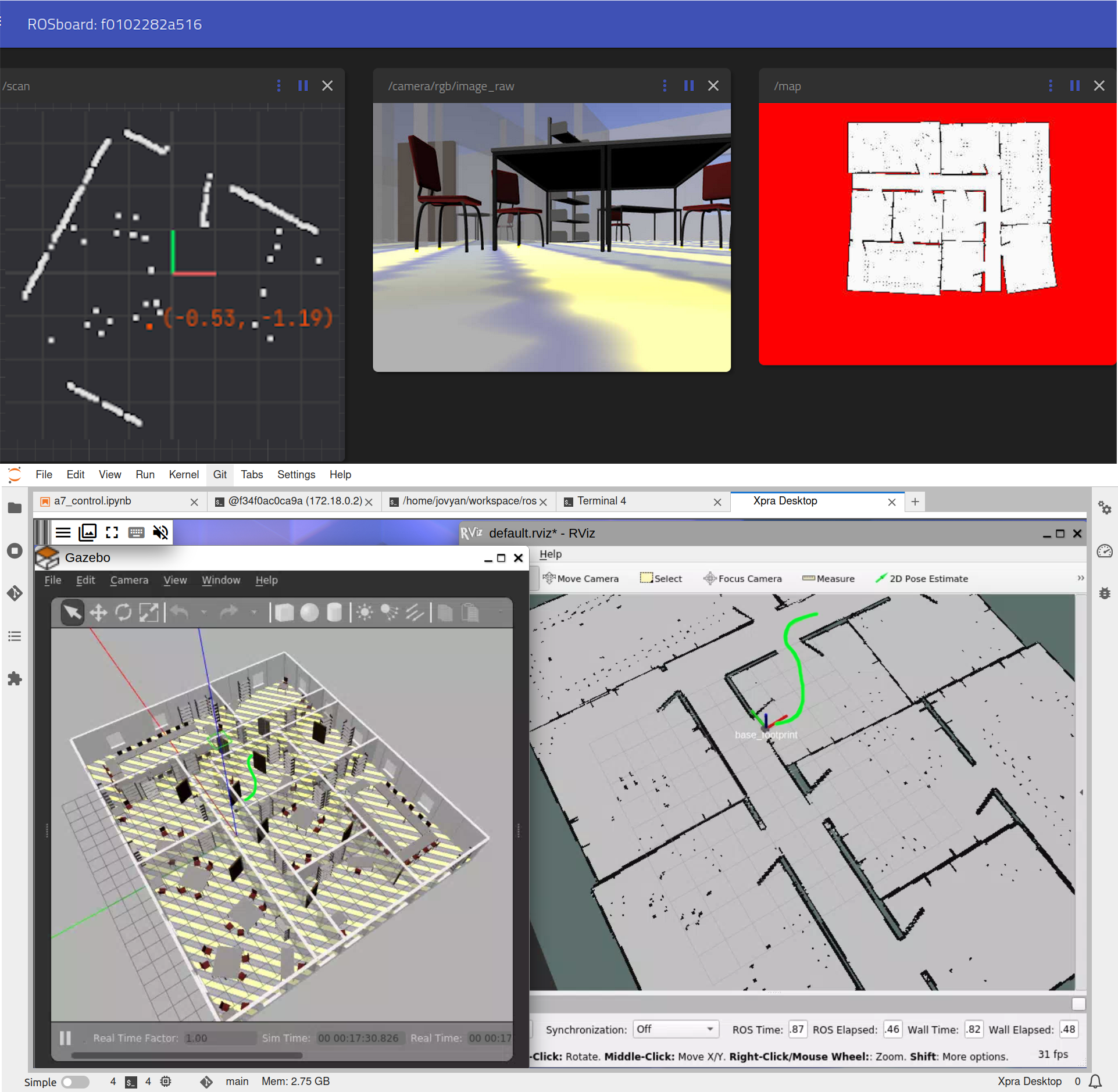}}
    \caption{ROSBoard \cite{JupyterROS} widget in Jupyter to visualize laser scan (top left), the front camera feed (top mid) and the recorded map (top right). Gazebo Physics Simulator (bottom left) and RViz visualization of the environment map (bottom right) on the XPRA VNC virtual desktop, navigating the Tortugabot through the \emph{digital twin} on the green path.}
    \label{fig:gazebo_vnc}
\end{figure}
    
    \paragraph{Into the Cloud}
    
        
    
        The pre-built containerized application is ready to run locally as well as to deploy on the cloud server. Our chosen cloud architecture is from the open-source project BinderHub, an open source web service that lets users create sharable, interactive, reproducible environments in the cloud \cite{ragan2018binder}. The underlying container orchestration system is Kubernetes, the de facto standard for deploying and operating containerized applications. Therefore, this setup can runs on major cloud platforms and self-hosted Linux servers. Additionally, it provides way to adjust computational resources without disrupting running services by adding or removing machines from the cluster.
    
        BinderHub is a platform designed for open-source workflows. To create a robotics containerized application, users only need to add an environment configuration file, usually a "Dockerfile," to their publicly available git repository. BinderHub builds a docker image accordingly and returns a URL of the application. This is equivalent to deploying a Software-as-a-Service (SaaS). Moreover, whenever there are updates in the code repository, BinderHub automatically rebuilds the application. Users can also build applications for different git branches or specific commits, thereby releasing different versions. With returned URL, anyone can use the published containerized application in a web browser. Each click on the URL creates a separate docker container for the current user.
        
        Figure \ref{fig:arch-2} illustrates the usecase of creating a virtual classroom using this cloud platform. The lecturer initially creates a GitHub repository to store learning materials and a Dockerfile of the runtime environment. Submitting the repository address to BinderHub's home page generates a URL for the virtual classroom. Simultaneously, the cloud server builds a docker image in the background. Once the building is complete, the lecturer distrubute the URL to the student. Hence learners can start the course immediately without installing any special software.
        

    \paragraph{Visualization}
        Visualization is imperative when teaching robotics to enable learners' volition through visual feedback. Live visualization of robot simulation in the web browser is not trivial. To visualize a \emph{digital twin}, a PC shows a 3D simulation on the display. A website can not simply visualize these programs, but there are ways to make it happen.
    
        The most popular visualization software in ROS is Rviz, it shows camera images, depth clouds, lasers, and most importantly, the current configuration of the robot. Jupyter is extended with RvizWeb, a plugin that reimplements Rviz for Jupyter and integrates into the web environment. ROSBoard is another plugin to Jupyter that provides visualization for specific ROS datastructures, like real-time laser scans, camera feeds and ground maps, as depicted in figure \ref{fig:gazebo_vnc} in the top. 
        
        A later addition to the Jupyter platform is a virtual display, such that programs with a graphical user interface can be shown in the browser. For physics simulation ROS offers Gazebo. In figure \ref{fig:gazebo_vnc} both Gazebo and Rviz are running in the online Docker container, visualized to the XPRA display via x-forwarding.




\section{Application}
\label{cha:application}

    With the opportunity to dive directly into a complicated architecture, it is possible to teach how to control robots, visualize sensor data and learn about general AI techniques in robotics. As the foundational platform we have used this technology to teach robotics at the Fall School \cite{FallSchool2023} since 2021, at IROS 2023 \cite{IROS2023Workshop} and in academic courses.

    \paragraph{Knowledge Representation and Reasoning}

        In our research, we have evaluated the usage of our system in teaching Knowledge Representation and Reasoning (KRR) for cognitive robots. This evaluation was conducted through its implementation in various university courses and Fall Schools. Our platform allows users to interact directly with the system using a knowledge representation querying language. One example of a KRR software we were teaching is KnowRob\cite{bessler2018}. KnowRob is a knowledge processing software for cognitive robots, which utilizes a Prolog-like language for queries. To effectively teach this querying language, we integrated our notebooks into courses such as "KI basierte Robotersteuerung" ("Ai-based robotic control") in 2022 and 2023.
        
    \paragraph{Knowledge Acquisition and Retrieval}

        In addition to a basic KRR course to teach how knowledge can be represented for the use in cognitive robots, we have also created a course on Web Knowledge Acquisition and its Retrieval by robotic agents that has been implemented in Fall Schools and will be extended and implemented in Workshop Tutorials. The semDT platform allows users to directly access Web content and use it to extend the robot knowledge base. The course also shows how robotic agents can access the knowledge base and reason about the contained knowledge. The notebooks will be integrated into courses in a new AI masters program.

    \paragraph{Task-Executives}
    
        Separate ROS applications can implement the functionality for navigation, image processing, knowledge representation and motion planning. The Cognitive Robot Abstract Machine (CRAM)\cite{beetz2023cram}
        is given a goal, like preparing a breakfast table and cleaning it up afterwards \cite{kazhoyan21easemilestone}.
        From this rather general description of a goal, the task executive gathers information about the environment, uses image processing to identify objects and queries the knowledge base for how to pick up the desired object, and navigating to the breakfast table, to finally acquire how and where to place it down without hitting any other items on the table. 
        Recurring lectures on Task Executives are found at the Fall School \cite{FallSchool2023} and the graduate course "Robot Programming with ROS" \cite{RPWR23}.
    


    \paragraph{Discussion}

        Prior to the cloud-based technology we offered VirtualBox \cite{Virtualbox} virtual machines (VM) with a full desktop environment, ROS and AI applications already installed. They were too large to download quickly, but served the broadest toolset for the individual student. Since we use RvizWeb, XPRA and ROSBoard we have visualization of the Docker image, so it can run in the web. Kubernetes was introduced to manage multiple clients. Running simulations requires a lot of computation time: we guarantee 8 GB of RAM per client session, more is needed for heavier digital twins, e.g. an Unreal Engine. 

        While hands-on tutorials with VirtualBox were unfeasible due to the VM size, offering the web platform, our proposed architecture, kept most attendees participating. At the Fall School 2023 we recorded the highest ratio in participation to attendance over the past years, thanks to the ease of access.

    \paragraph{Future Work}

        To enable interested learners to get into the field of cognitive robotics we are creating courses and tutorials. While our focus is currently on higher education and in-service training, secondary education is planned.

        While our server runs Kubernetes to orchestrate new students launching the digital-twin environment, we were looking into two other platforms that provide computational resources for open-source software. Github Codespaces \cite{GithubCodespaces} fully supports running our robot application, but the cloud computing resources are strictly limited. Google Colab \cite{GoogleColab} provides robust computational resources, including GPUs, yet imposes restrictions on the WebSocket protocol, which renders our visualization tool dysfunctional.
        
\section{Conclusion}
\label{cha:conclusion}


    Building containerized applications is critical for cloud-based robot systems and is a fundamental task of cloud solutions in many other research domains. Within this setup, a significant challenge is the real-time rendering of 3D robot digital twin environments and displaying them on a web-based client to achieve platform independent accessibility. 
    
    Such a cloud-based and independent digital twin learning \& training platforms solves the aforementioned setup and hardware problems. 

    Tech giants like Amazon and NVIDIA have launched their robot simulation cloud services, providing much more computational resources and cutting-edge technologies. However, our focus remains on free and open-source solutions tailored for the education and research domains, and to encourage more researchers, developers, and educators to engage in robotics.

\bibliographystyle{IEEEtran}
\bibliography{24-EDUCON-Cognitive-Robots-in-the-Cloud}

\end{document}